\documentclass[12pt]{article}
\usepackage[margin=.8in]{geometry}
\usepackage{authblk} 
\usepackage{comment} 
\usepackage{fullpage} 
\usepackage{xcolor}
\usepackage{subfiles}

\usepackage{array}
\usepackage{booktabs}
\usepackage{xurl} 
\usepackage{hyperref}
\hypersetup{breaklinks=true, colorlinks=true, urlcolor=blue,citecolor=green} 
\usepackage{algpseudocode}

\usepackage{times}
\usepackage{mathtools}
\usepackage{amsmath}
\usepackage{bm}
\usepackage{amssymb}
\usepackage{pifont}
\usepackage{marginnote}
\usepackage{graphicx}
\usepackage{dcolumn}%
\usepackage{dcolumn}
\usepackage{bm}
\usepackage{mathtools}
\usepackage{color}
\usepackage[linesnumbered,ruled,vlined]{algorithm2e}

\usepackage[font=small,labelfont=bf]{caption}

\makeatletter
\renewcommand{\maketitle}{\bgroup\setlength{\parindent}{0pt}
\begin{flushleft}
 \fontsize{16}{32}
 \textbf{\@title}
 
   \fontsize{12}{24}
  \@author
\end{flushleft}\egroup
}
\makeatother
\newfam\bboardfam
\font\tenbboard=msbm10  
 \font\sevenbboard=msbm7
   \font\fivebboard=msbm5 
\textfont\bboardfam=\tenbboard
\scriptfont\bboardfam=\sevenbboard
\scriptscriptfont\bboardfam=\fivebboard

\title{Multi-variable Adversarial Time-Series Forecast Model}

\author[1]{Xiaoqiao Chen}

\affil[1]{Department of Computing and Mathematical Sciences, California Institute of Technology. Pasadena, California, 91125, USA.}

\date{June 2021}

\begin{document}

\maketitle
\section*{Abstract}
Short-term industrial enterprises power system forecasting is an important issue for both load control and machine protection. Scientists focus on load forecasting but ignore other valuable electric-meters which should provide guidance of power system protection. We propose a new framework, multi-variable adversarial time-series forecasting model, which regularizes Long Short-term Memory (LSTM) models via an adversarial process. The novel model forecasts all variables (may in different type, such as continue variables, category variables, etc.) in power system at the same time and helps trade-off process between forecasting accuracy of single variable and variable-variable relations. Experiments demonstrate the potential of the framework through qualitative and quantitative evaluation of the generated samples. The predict results of electricity consumption of industrial enterprises by multi-variable adversarial time-series forecasting model show that the proposed approach is able to achieve better prediction accuracy. We also applied this model to real industrial enterprises power system data we gathered from several large industrial enterprises via advanced power monitors, and got impressed forecasting results.

\section*{Introduction}
Short-term power system forecasting are wildly used by power sectors, smart grid designer, and other electricity sectors for periodical operations. Always, load forecasting studies focus on aggregated load in a large area, such as a country, a state, or a city. But recent years, along with fast growing electricity consumption and expensive electricity cost brought by it, the demands of electric load forecasting for a single individual, especially modern large industrial enterprises, are sharply increasing but seldom receive much attention from scientists.
Moreover, besides power load, other electric-meters like voltage, current, power, power factor, etc. are also essential for both firms themselves and power system. For enterprises themselves, machines are suggested to run under right voltage or current, otherwise the bad situation will accelerate machine aging. For power system, most power grid can not deal with suddenly increased power pressure. With accurate electric-meter forecasting, large firms can maintain power system stable through off-peak production activities to avoid fluctuation. 

There are two main challenges in this problem: data gathering and accurate forecasting model. Normally, firms don’t equip professional measurement instruments for detailed, sub-area and short-term electric power parameters. Therefore, it is hard to get real data for forecasting. Moreover, even though detailed data is available, the single industrial enterprise short-term power system forecasting is still the most challenging in model and framework level. Compared with large-scale load forecasting, customer-wise forecasts are conventionally considered trivial due to high volatility and uncertainty of individual loads. Furthermore, compared with power load forecasting, we’re expected to forecast batch of variables at the same time. It is harder to make it accurate for each single variable. Therefore, we can not expect good performance if apply on-the-shelf load forecasting models to industrial enterprise scenario even if they have been proven successful in large-scale load forecasting problems. A novel forecasting framework, which aims to individual firm multi-variable power system, is needed for accurate forecasting. 
Abstracting the power system problem into a supervised sequence-to-sequence problem, that is multi-variable time-series forecast problem. The goal of multi-variable time-series forecast problem is to forecast a batch of dependent variables at time t based on variables before time t-1 in supervised manner.

Recently, researches have proposed a number of schemes and techniques for the supervised time-series sequence forecast, which can be divided into two main categories: a single model with multi-variables output and multi-models with one variable output for each model. 

The first-type methods, as the model views variables as a whole, considers the dependence between variables. The architecture is shown in Figure 1.
\begin{figure}
  \centering
  \includegraphics[width=1\textwidth]{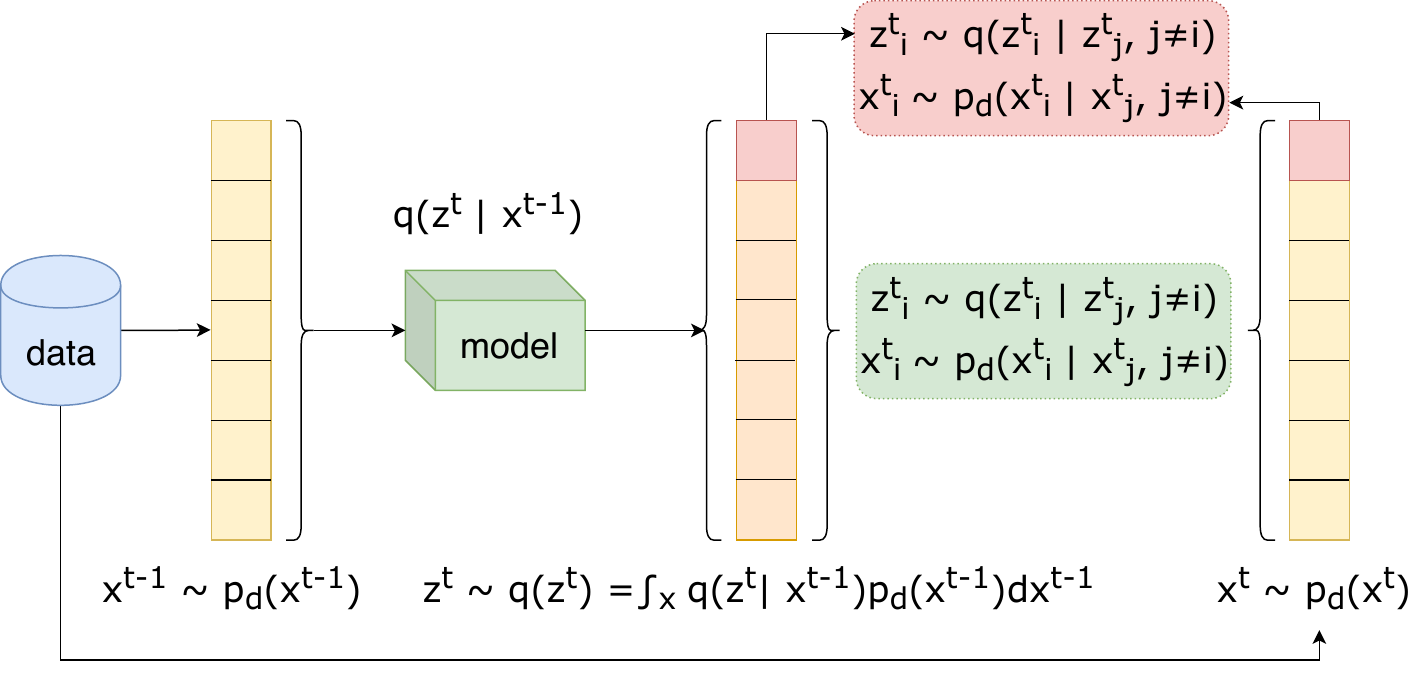}
  \caption{Standard multi-output forecasting model}
\end{figure}

This kind of models have some disadvantages as follows: 

1.	A single model can’t deal with different types of variables (continue variables, category variables, integer variables, etc.) together and get high accuracy for all of them. 

2.	The single model minimizes the loss function as an integrated loss among all variables, which is more like a “sum”. If we pick up one variable, it may not in its optimal position. That is to say, it’s hard to get forecast error for each variable as low as forecasting it separately.

3.	If the model is a multi-layer neural network, considering each variable and its corresponding network part separately, normally the difference only lies on the last layer while all previous layers are the same. It is never a good idea to share parameters for outputs in different types and with different meanings, which highly weakens the network.

The second-type methods, forecast each variable separately so that we get optimal solution for each variable and it’s flexible for different variable types. The disadvantages are primarily that it never consider the dependence between variables. The architecture is shown in Figure 2.
\begin{figure}
  \centering
  \includegraphics[width=1\textwidth]{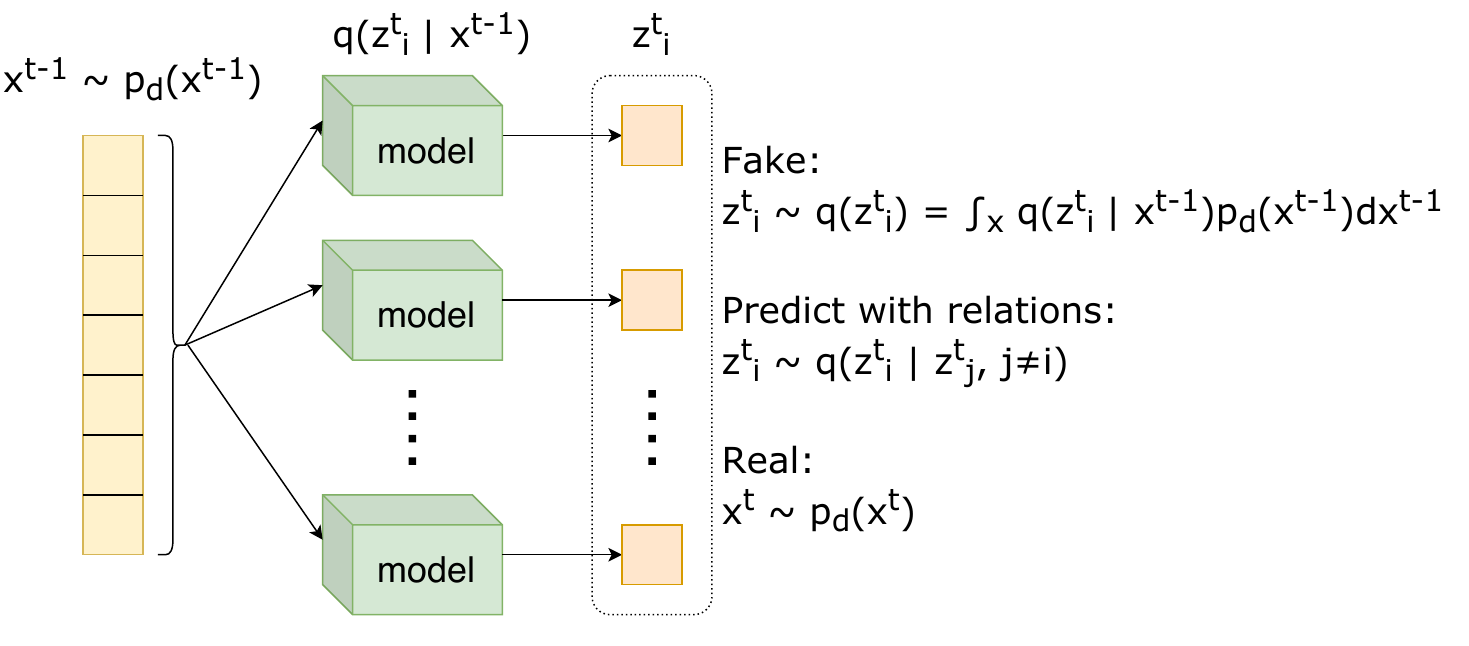}
  \caption{Parallel single-output forecasting models}
\end{figure}

In summary, the challenge is to overcome the problem of inaccurate prediction of individual variables when the overall loss function is minimized, and to consider the correlation between variables when modeling each variable separately. 

For improvement, we present a hybrid framework which the adversarial network and the single-variable time-series forecast models are trained jointly with SGD in two phases – the forecast phase and the regularization phase – executed on each mini-batch.  

\section*{Related Work}

\subsection*{Power System Forecasting}
Most short-term load forecasting studies focus on large-scale power system or substation level. The most popular and widely used in industry is autoregressive integrated moving average (ARIMA) model, which is first proposed by Cao et al. $^{[1]}$ ARIMA has good performance in intraday load forecasting. Another popular method for shorter-term forecasting is radial basis function (RBF) neural network introduced by Z. Yun et al. in 2018.$^{[2]}$ For more efficient forecasting, k-nearest neighbour (KNN)$^{[3]}$ algorithm has been proven successful in many experiments and simulations.$^{[4]}$ 

Recently, some scientists prefered to use LSTM neural network for load forecasting. LSTM neural network, a type of recurrent neural network (RNN), which was originally introduced by Hochreiter and Schmidhuber$^{[5]}$, has received enormous attention in the sequence learning scenario, including load forecasting. D. L. Marino, K. Amarasinghe and M. Manic$^{[6]}$ proposed a standard LSTM framework and a Sequence to sequence (S2S) architecture LSTM framework to predict building level power load. They compared the two framework and showed that the network with the S2S architecture outperformed the standard LSTM one. After that, Tomas Jarabek et al.$^{[7]}$ combined the two LSTM framework with k-shape$^{[8]}$ clustering and got promoting performance.

A few of scientists have confronted with individual customers directly. Ghofrani et al.$^{[9]}$ is the first scientist that focus on load forecasting for individual users in 2011. After that, Chaouch$^{[10]}$ proposed a clustering-based forecasting method for intra-day household-level load in 2014. In 2015, P. Zhang et al.$^{[11]}$ developed a big data architecture of decision tree to predict power load based on smart meter data. Most recently in 2019, a customer-wise short-term load forecast framework based on LSTM is introduced by Weicong Kong et al.$^{[12]}$, which is the first example of individual power load forecasting based on LSTM. 

These a few previous studies on customer-wise load forecasting are all about household users, which are different from industrial enterprises. And most of these models (including the LSTM based method) apply state-of-art large-scale load forecasting models to customers-wise problems directly or with only fine adjustment. Therefore, the issues remains open for more accurate results.
\subsection*{Multi-variable time-series forecasting}
The promise of deep learning is to discover rich, hierarchical models[Generative Adversarial Nets]. So far, the most striking successes in deep learning have involved discriminative models, usually those that map a high-dimensional, rich sensory input to a class label $^{[13,14]}$. Deep generative models have had less of an impact, due to the difficulty of approximating many intractable probabilistic computations that arise in maximum likelihood estimation and related strategies, and due to difficulty of leveraging the benefits of piecewise linear units in the generative context. 

Ian J. Goodfellow proposed a new generative model call Generative Adversarial Nets (GAN)$^{[15]}$ estimation procedure that sidesteps these difficulties. Generative adversarial networks are an example of generative models. In the proposed adversarial nets framework, the generative model is pitted against an adversary: a discriminative model that learns to determine whether a sample is from the model distribution or the data distribution. Competition in this game drives both teams to improve their methods until the counterfeits are indistiguishable from the genuine articles.

The advantages of GAN are that Markov chains are never needed, only backprop is used to obtain gradients, no inference is needed during learning, and a wide variety of functions can be incorporated into the model. GANs are generative models that use supervised learning to approximate an intractable cost function, much as Boltzmann machines use Markov chains to approximate their cost and VAEs use the variational lower bound to approximate their cost. GANs can use this supervised ratio estimation technique to approximate many cost functions, including the KL divergence used for maximum likelihood estimation.$^{[16]}$ 

GANs are relatively new and still require some research to reach their new potential. Researchers should strive to develop better theoretical understanding and better training algorithms. GANs are crucial to many different state of the art image generation and manipulation systems, and have the potential to enable many other applications in the future.

 A feedback network called "Long Short-Term Memory"$^{[17]}$ overcomes the fundamental problems of traditional RNNs, and efficiently learns to solve many previously unlearnable tasks. The basic unit in the hidden layer of an LSTM network is the memory block;$^{[18]}$ it replaces the hidden units a “traditional” RNN. A memory block contains one or more memory cells and a pair of adaptive, multiplicative gating units with gate input and output to all cells in the block. Memory blocks allow cells to share the same gates thus reducing the number of adaptive parameters. LSTM has transformed machine learning and Artificial Intelligence (AI)$^{[19]}$, and is now available to billions of users through the world's four most valuable public companies$^{[20]}$: Apple, Google (Alphabet), Microsoft, and Amazon. Over the past decade, LSTM has proved successful at a range of synthetic tasks requiring long range memory and various real-world problems$^{[21]}$, such as learning context free languages$^{[22]}$, protein secondary structure prediction.$^{[23,24]}$ As would be expected, its advantages are most pronounced for problems requiring the use of long range contextual information. This characteristic makes it possible for LSTM to apply to load forecasting in power systems.

\section*{Method}
The state-of-art time-series models aren’t able to optimize each variable and consider the relations between them at the same time. We proposed the multi-variable adversarial time-series forecast model help to trade off the accuracy of each separated variable and the regularization relations between variables. 

Let $x^{t-1}$ be the input and $z^t$ be the forecast output of the LSTM model. $x^{t-1}$ is some shape of data abstracted from the known data from time $0$ to time $t-1$, for example the last week data or the last month data. $z^t$ is the integrated output from forecasting models. And $x^t$ is some shape of target data at or after time $t$. Also let $p_d(x)$ be the raw data distribution. 

The multi-variable adversarial time-series forecast model consists of a batch of single-variable time-series forecasting models $M_i(x)$, where the integrated forecast output is regularized by matching the concatenated output, $q(z^t)$, to the raw data with relations between variables, $p_d(x^t)$. In order to do so, an adversarial network is attached on the concatenation results of all forecast model outputs as illustrated in the following figure, while the LSTM forecast models try to get accurate forecast to confuse the discriminator. The whole model is in adversarial manner. The architecture of this novel model is shown in Figure 3.

\begin{figure}
  \centering
  \includegraphics[width=1\textwidth]{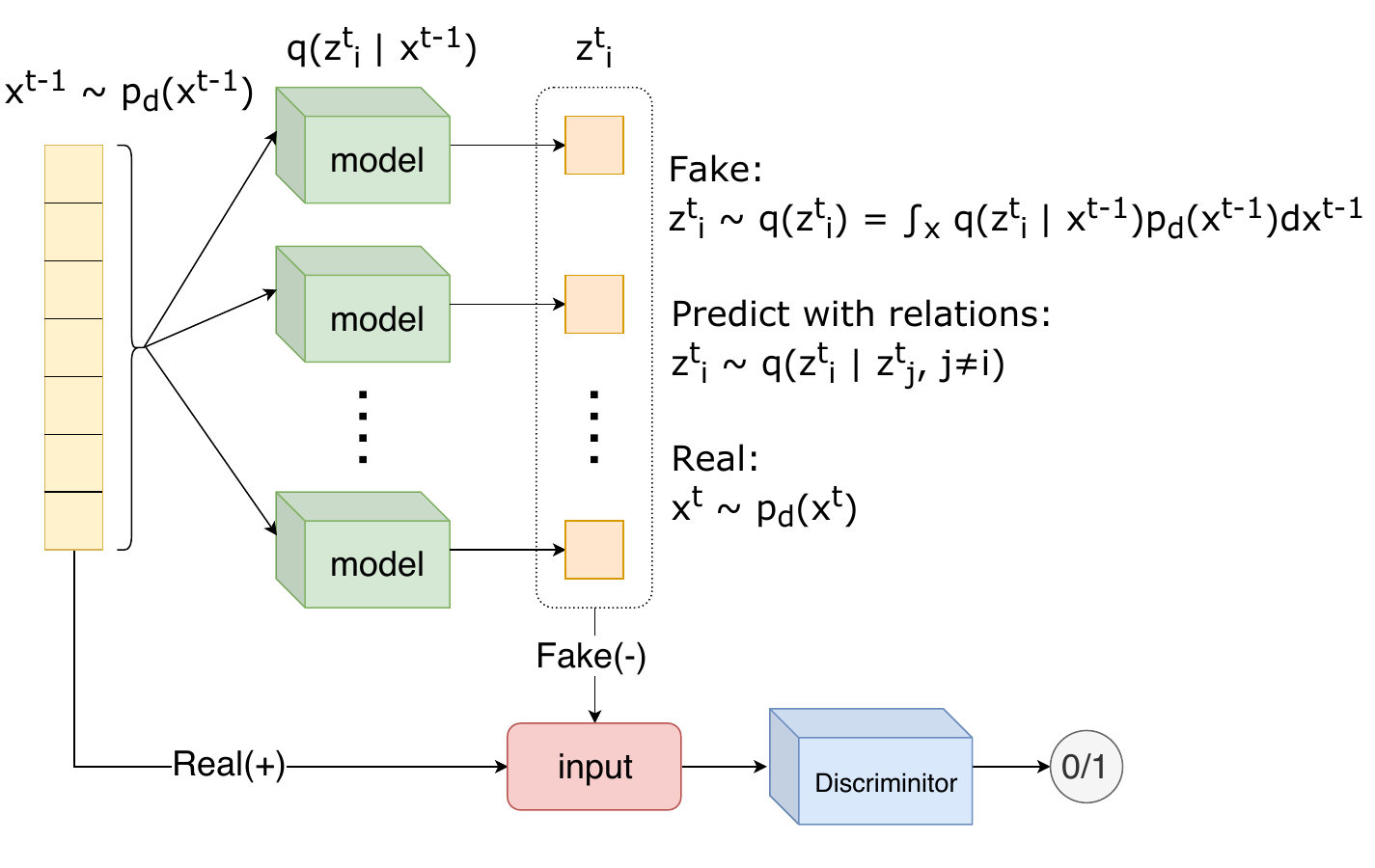}
  \caption{Multi-variable adversarial time-series forecasting model}
\end{figure}

Each single-variable time-series forecasting model, with distribution $q(z^{t}_i|x^{t-1})$ focuses on forecasting one variable $z^{t}_i$ and defines an independent forecast result of $z^{t}_i$, $q(z^{t}_i)$ as follow:
\begin{equation}
    z^{t}_i \sim  q(z^{t}_i) = \int_{x^{t-1}}q(z^{t}_i|x^{t-1})p_d(x^{t-1})dx^{t-1}
\end{equation}

Which is the distribution without variable-variable relations (not conditional distribution). We try to guide $q(z^{t}_i)$ to $q(z^{t}_i|z^{t}_j, j\ne i)$, the distribution with variable-variable relations (conditional distribution). The generator of the adversarial network is also the forecast models $q(z^{t}_i|x^{t-1})$. The forecast models attempts to minimize the forecast error to ensure the aggregated posterior distribution can fool the discriminative adversarial network into thinking that the concatenation output $q(z^t)$ comes from the true prior distribution $p_d(x^t)$. 

In the adversarial network, the discriminator model, $D(x)$, just like GANs, is a neural network that tries to discriminate that the input $x$ is a sample from the raw data distribution with variable-variable relations (positive samples) rather than a fake sample which is the concatenation of the outputs of single-variable forecasting models without variable-variable relations. 

The single-variable time-series forecast models and discriminator are trained jointly with SGD in training batch in two phases – the forecast phase and the regularization phase. 

In the forecast phase, each single-variable time-series forecast model trained to minimize the forecast error as follow:
\begin{equation}
    \min_{M_i} \frac{1}{n}(x^{t}_i - M_i(x^{t-1})^2
\end{equation}

In the regularization phase, the discriminator first updates its parameters to classify the real samples (from raw data) from the forecasted samples (the concatenation output from forecast models). The adversarial network then trains all single-variable forecast models to fool the discriminative network. The solution to this game can be expressed as follow:
\begin{equation}
    \min_{M_i} \max_{D} E_{x^t \sim p_d(x^t)}[\log D(x^t)]+E_{x^{t-1} \sim p_d(x^{t-1})}[\log (1-D(concat(M_i(x^{t-1}))))]
\end{equation}

\section*{Results}
\subsection*{Weather Forecasting}
I first worked on a standard time-series forecast dataset, Air Quality dataset, from UCI Machine Learning Repository$^{[5]}$ to test the multi-variable adversarial time-series forecast model.

This is a dataset that reports on the weather conditions of Beijing, China each hour for five years.

The data includes the date-time, the pollution called PM2.5 concentration, dew point, temperature, pressure, wind direction, wind speed and the cumulative number of hours of snow and rain. The complete feature list in the raw data is as follows:

1.	No: row number

2.	year: year of data in this row

3.	month: month of data in this row

4.	day: day of data in this row

5.	hour: hour of data in this row

6.	pm2.5: PM2.5 concentration

7.	DEWP: Dew Point

8.	TEMP: Temperature

9.	PRES: Pressure

10.	cbwd: Combined wind direction

11.	Iws: Cumulated wind speed

12.	Is: Cumulated hours of snow

13.	Ir: Cumulated hours of rain

The plot of raw data is shown in Figure 4.

\begin{figure}
  \centering
  \includegraphics[width=0.5\textwidth]{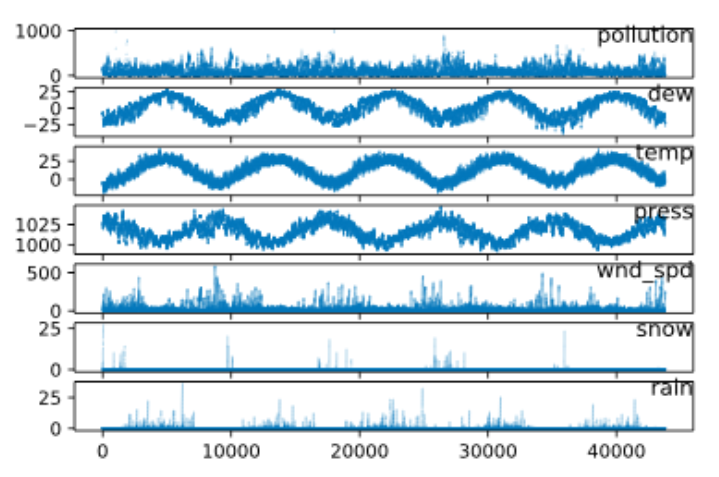}
  \caption{Air Quality dataset}
\end{figure}

Given the weather conditions for prior 24 hours, we forecast seven valuable features (pm2.5, DEWP, TEMP, PRES, Iws , Is, and Ir) at the next hour. We use 3 layers stacked LSTM model with 10 units in each layer as the single-variable forecast model.

The loss and accuracy of discriminator are shown in Figure 5.
\begin{figure}
  \centering
  \includegraphics[width=0.49\textwidth]{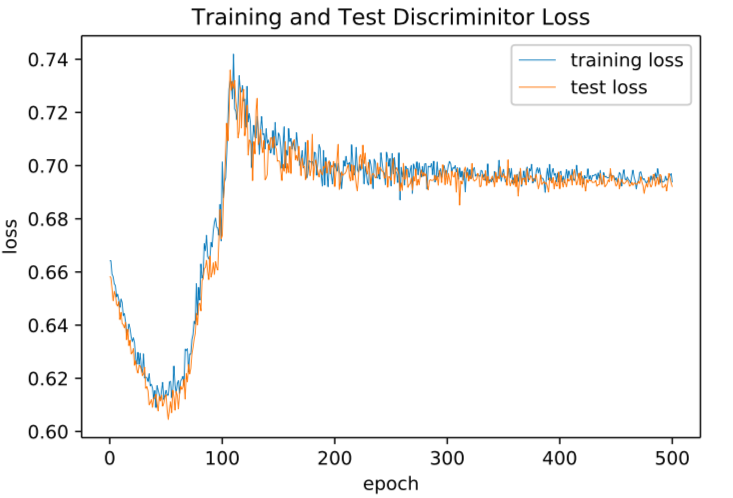}
  \includegraphics[width=0.49\textwidth]{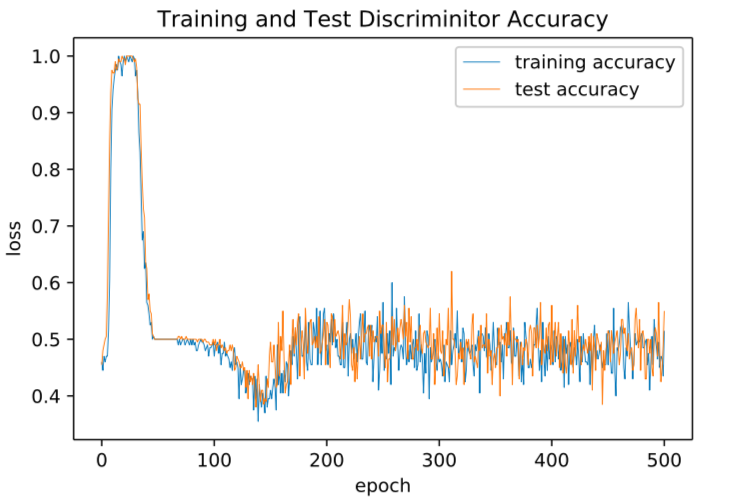}
  \caption{Loss and accuracy of discriminator}
\end{figure}

We get outstanding results in time-series forecasting, the training and test results are shown in Figure 6.
\begin{figure}
  \centering
  \includegraphics[width=0.49\textwidth]{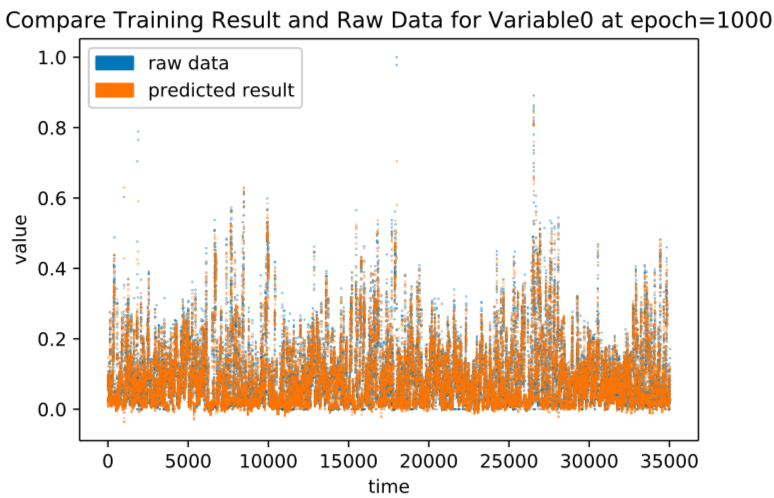}
  \includegraphics[width=0.49\textwidth]{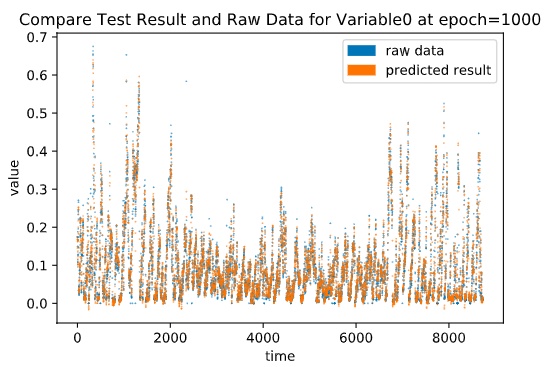}
  \includegraphics[width=0.49\textwidth]{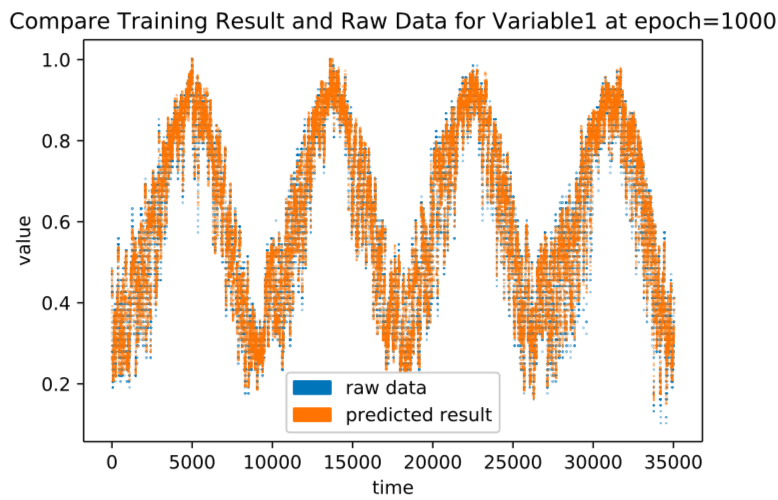}
  \includegraphics[width=0.49\textwidth]{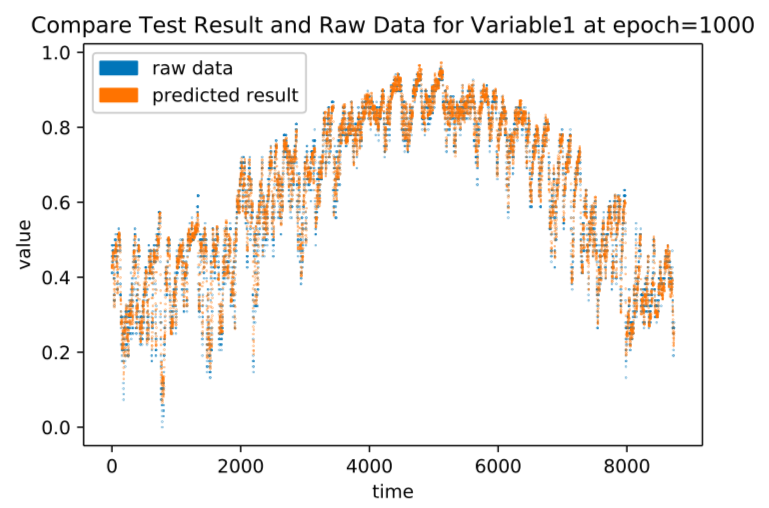}
  \includegraphics[width=0.49\textwidth]{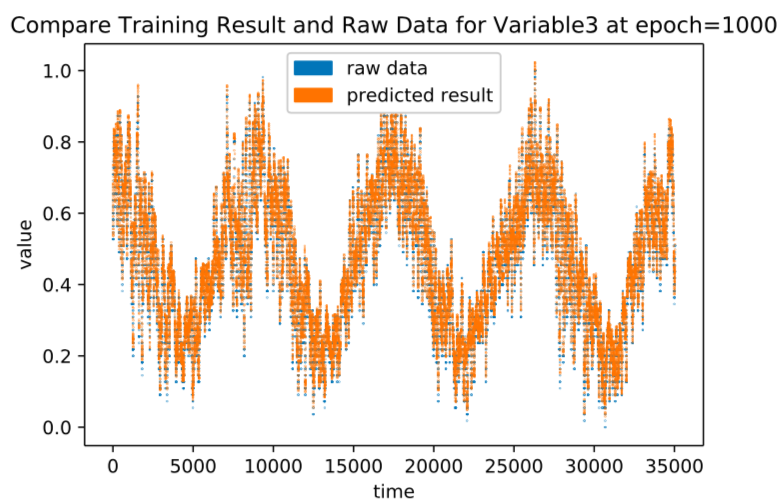}
  \includegraphics[width=0.49\textwidth]{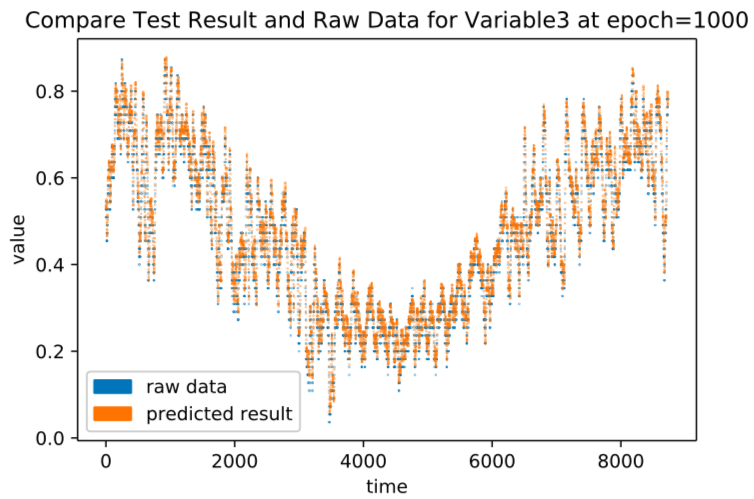}
  \includegraphics[width=0.49\textwidth]{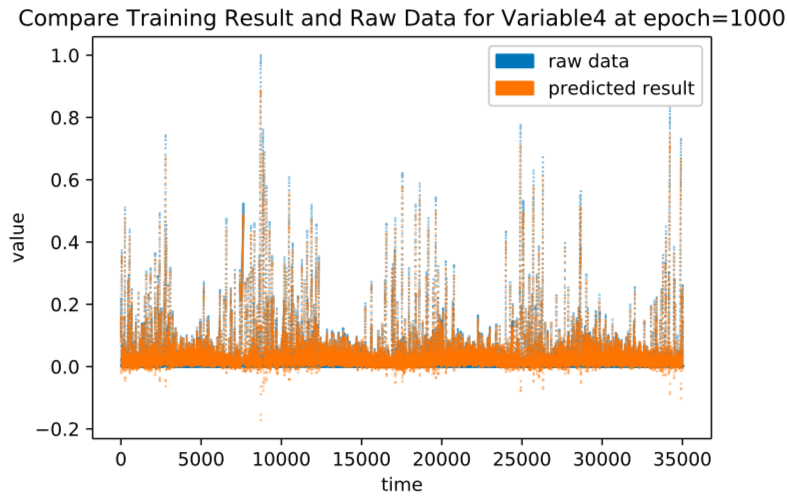}
  \includegraphics[width=0.49\textwidth]{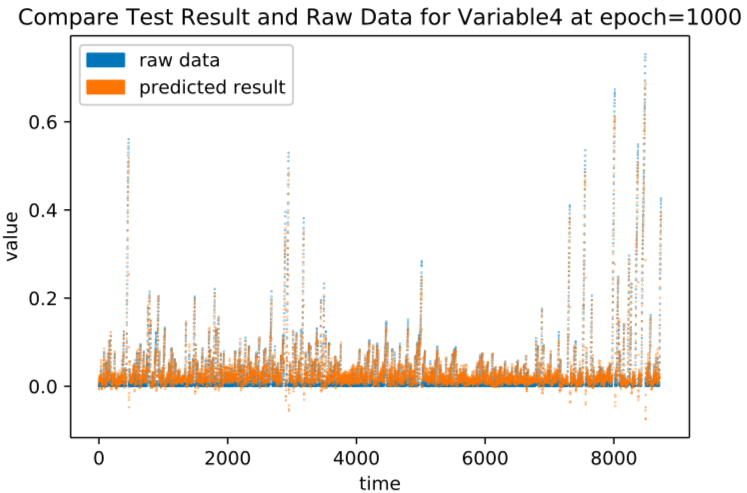}
  \includegraphics[width=0.49\textwidth]{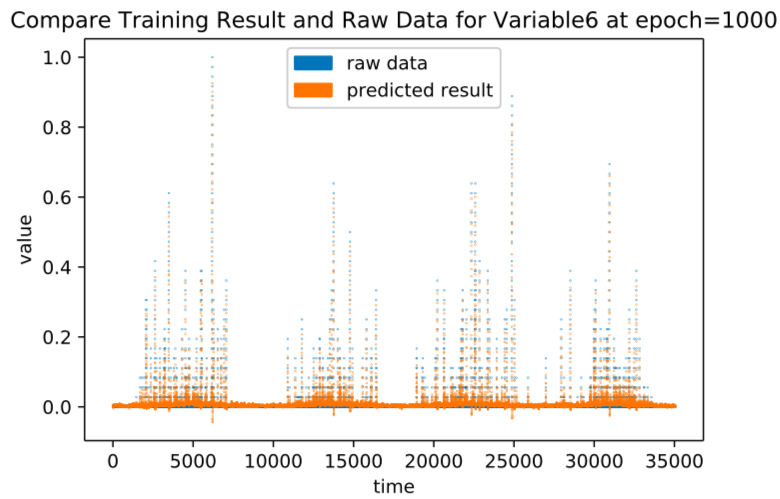}
  \includegraphics[width=0.49\textwidth]{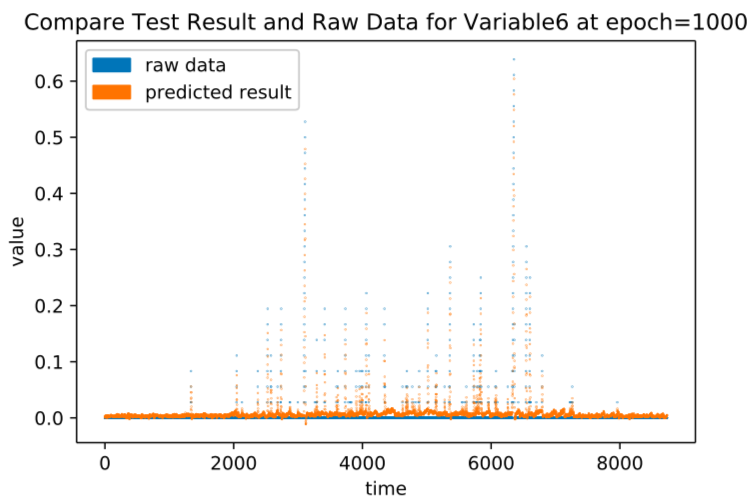}
  \caption{Forecasting results of Air Quality dataset}
\end{figure}

To compare the multi-variable adversarial time-series forecast model with standard LSTM model, we trained a standard LSTM networks with similar architecture (3 layers stacked LSTM model with 10 units in each layer) using the same dataset. The mean squared error of each variable forecasted by multi-variable adversarial time-series forecast model is better than standard LSTM model. The compared results are shown in Figure 7.
\begin{figure}
  \centering
  \includegraphics[width=0.49\textwidth]{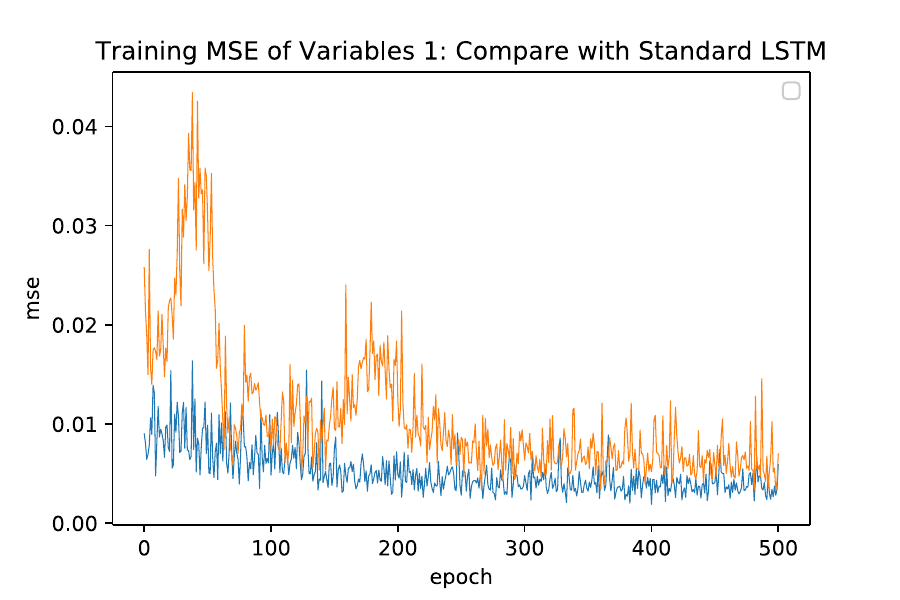}
  \includegraphics[width=0.49\textwidth]{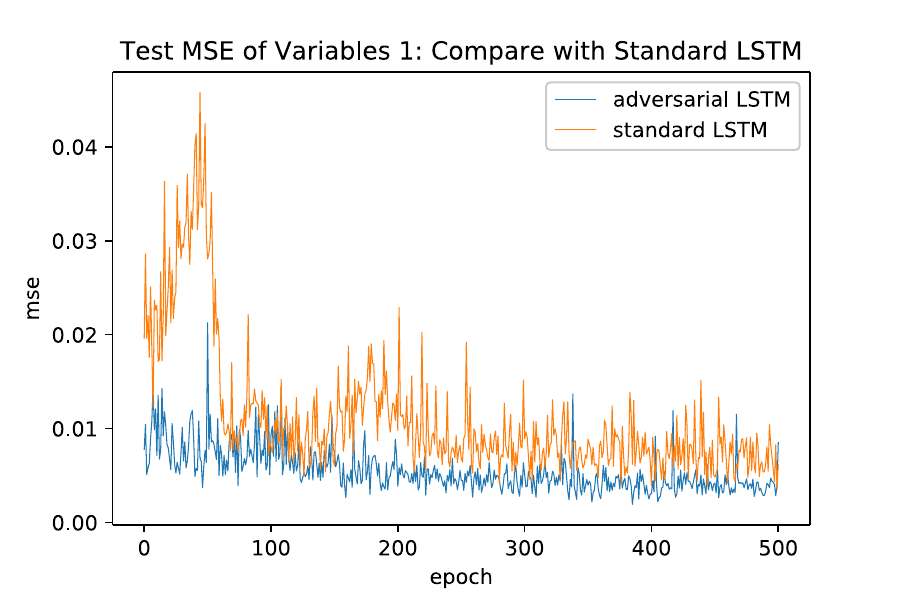}
  \includegraphics[width=0.49\textwidth]{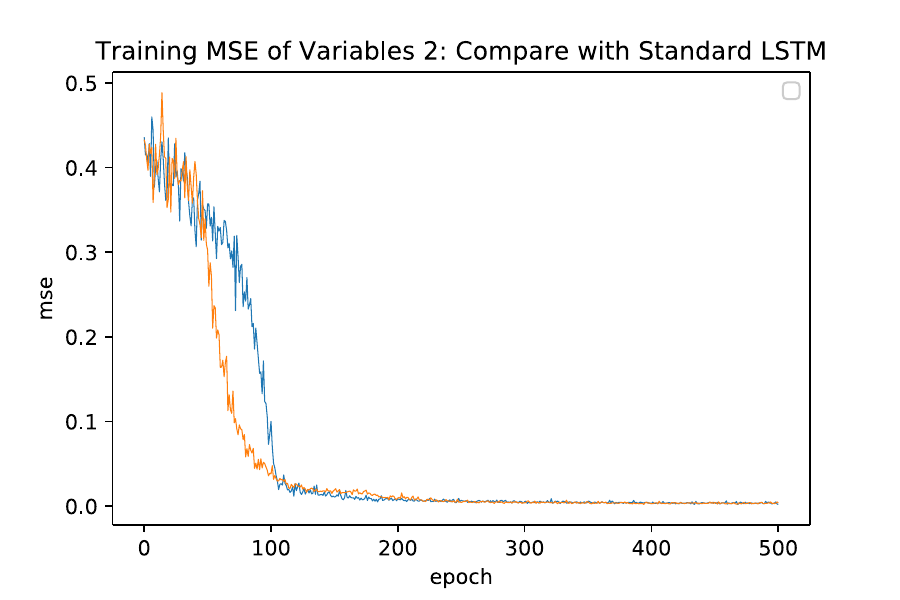}
  \includegraphics[width=0.49\textwidth]{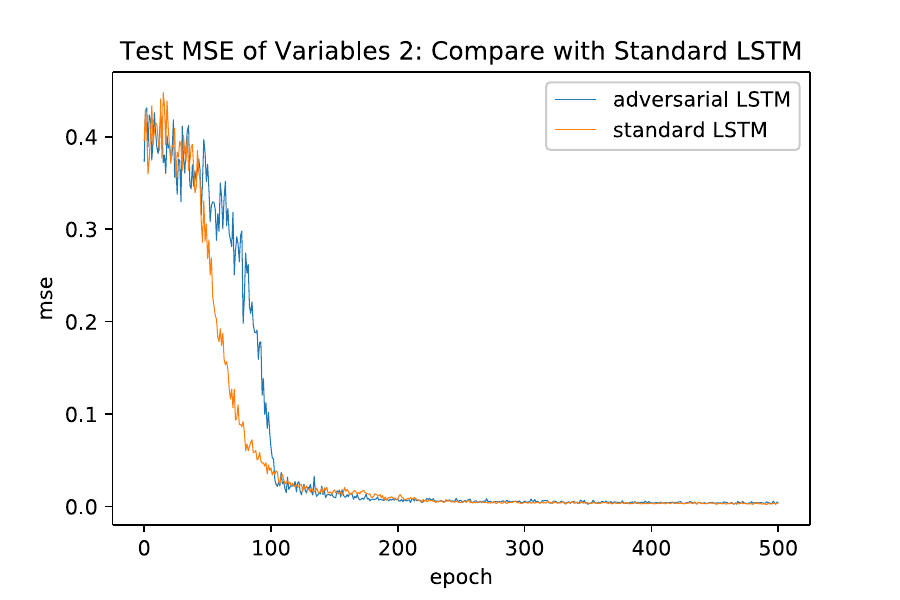}
  \includegraphics[width=0.49\textwidth]{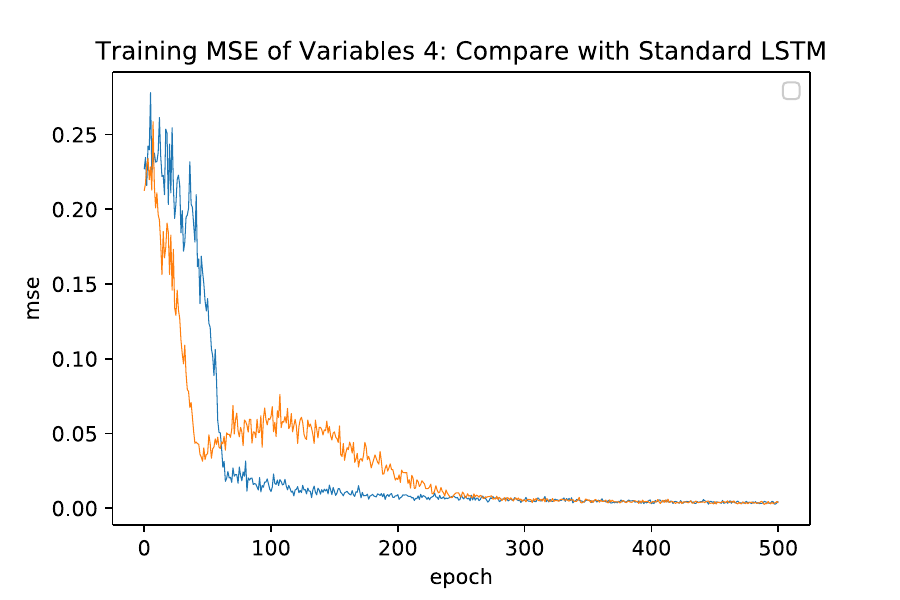}
  \includegraphics[width=0.49\textwidth]{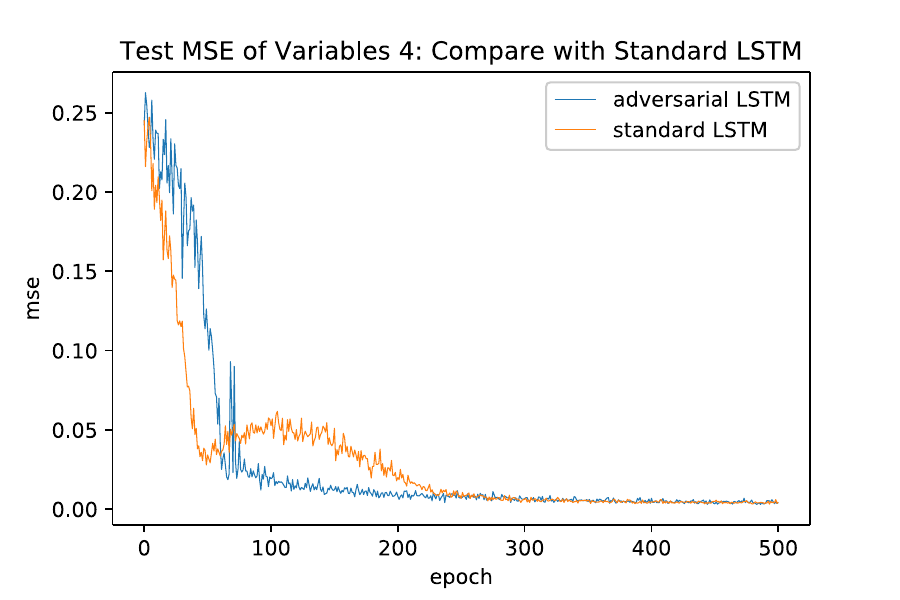}
  \includegraphics[width=0.49\textwidth]{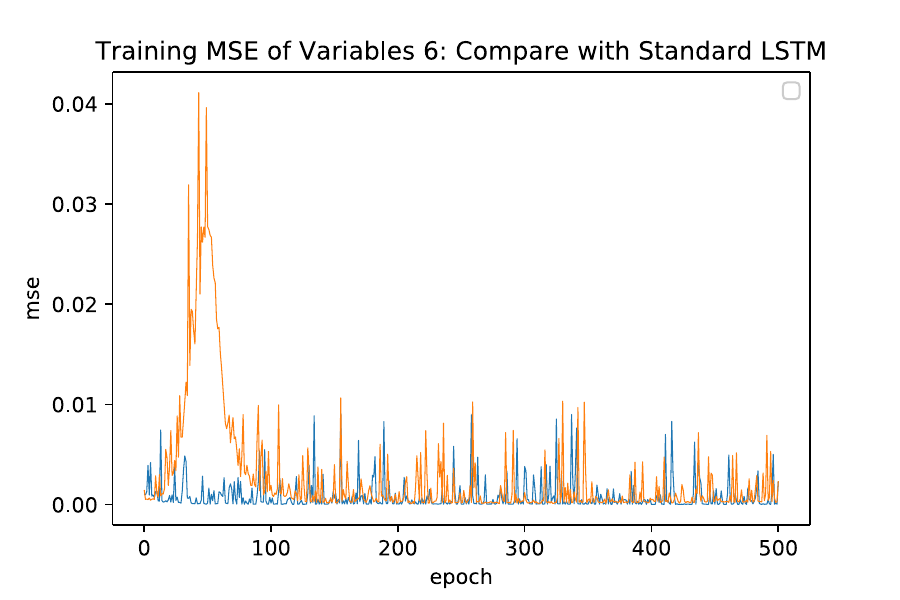}
  \includegraphics[width=0.49\textwidth]{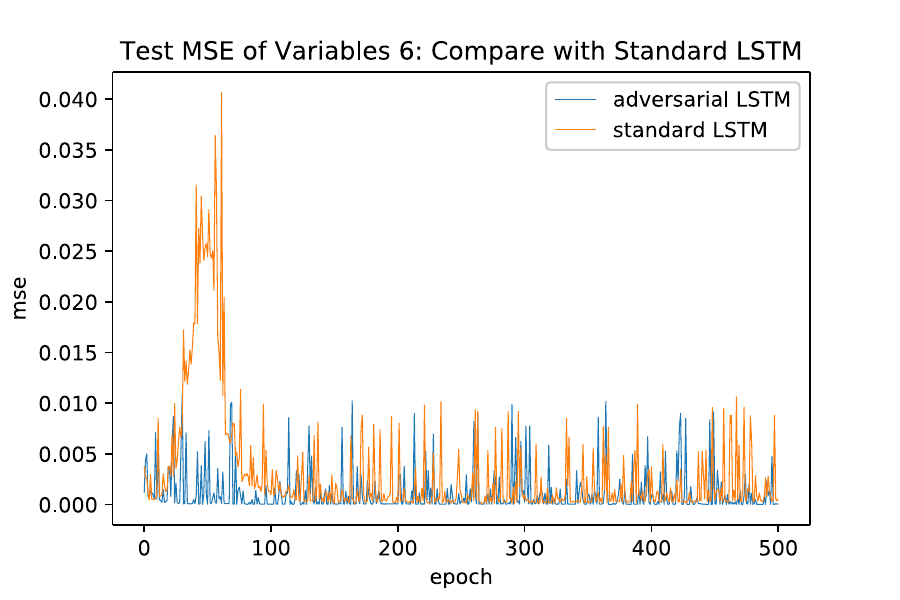}
  \includegraphics[width=0.49\textwidth]{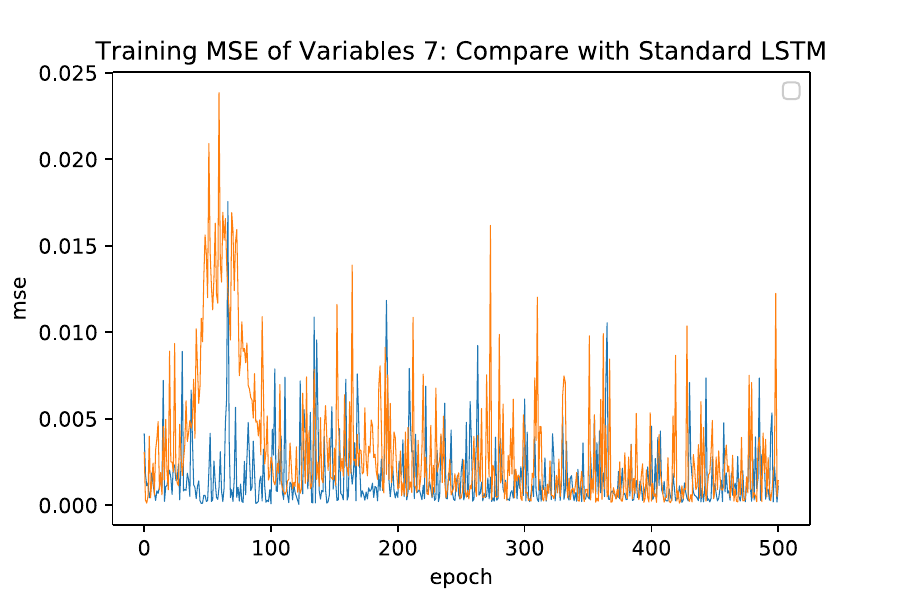}
  \includegraphics[width=0.49\textwidth]{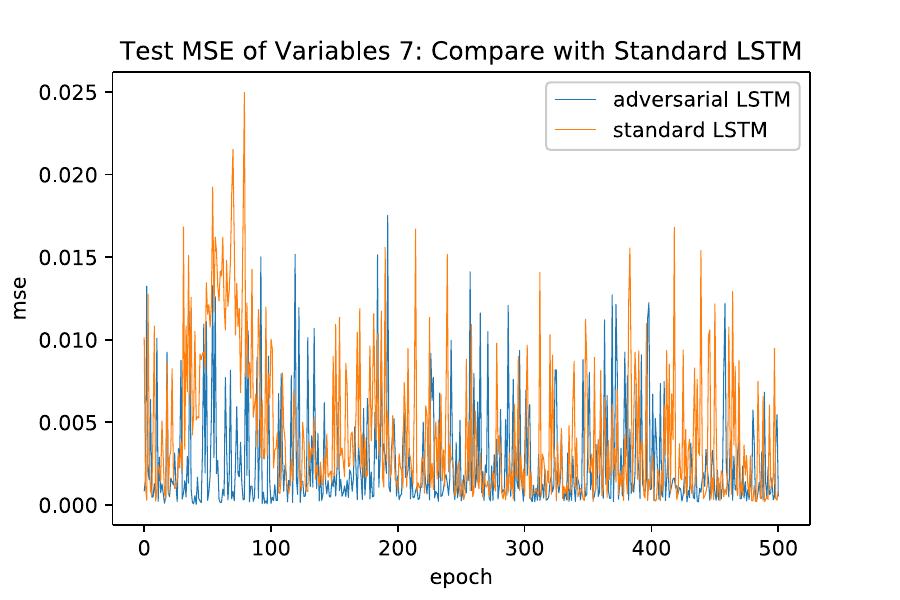}
  \caption{Mean squared error of each variable: compare with standard LSTM network}
\end{figure}

\subsection*{Industrial Enterprises Power System Data}

The data is from the big data industrial power load monitor system developed and owned by Changfu Power System Analysis and Design Ltd, which collaborates with us and provide us exclusively with constantly updated large data sets of power load data and related electric parameters gathered from dozens of large industrial enterprises (Ausnutria, etc). 

The data is gathered every 15 minutes from November 2017 to present for all sub-areas (such as sub-workshops) in each enterprise. The electric parameters are load, active power, reactive power, three-phase voltage, three-phase current, three-phase power, three-phase power factor, temperature, etc. The plot of part of this dataset is shown in Figure 8.

We can use this data and frame a forecasting problem where, given the electric parameters for prior day (96 points), we forecast all electric parameters at the next 15min. We use single layer LSTM model with 100 units as the single-variable forecast model.

Here is the forecast results in Figure 9. The predicting results fit the target well, which should be a reliable guidance for industrial enterprises to make decision for machine protection.

\begin{figure}
  \centering
  \includegraphics[width=0.49\textwidth]{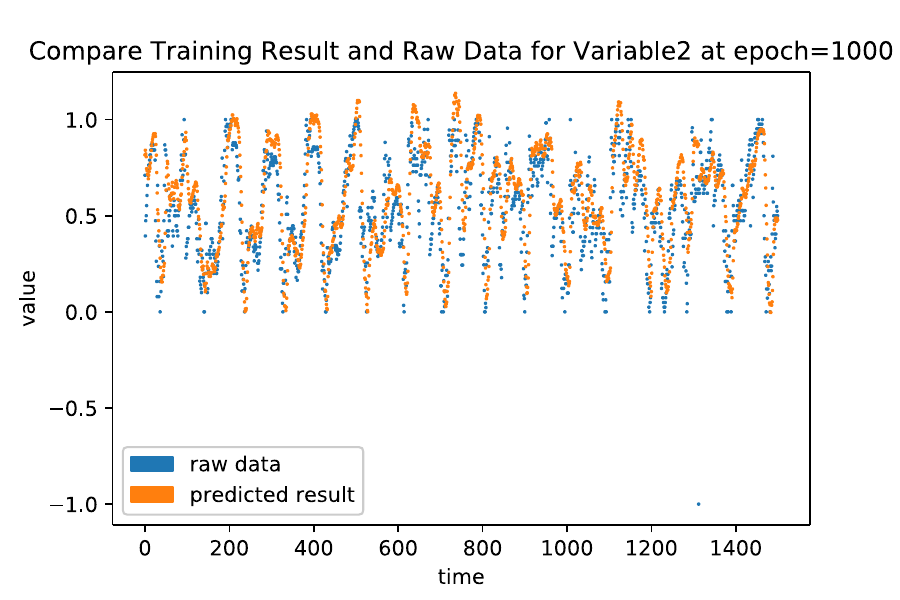}
  \includegraphics[width=0.49\textwidth]{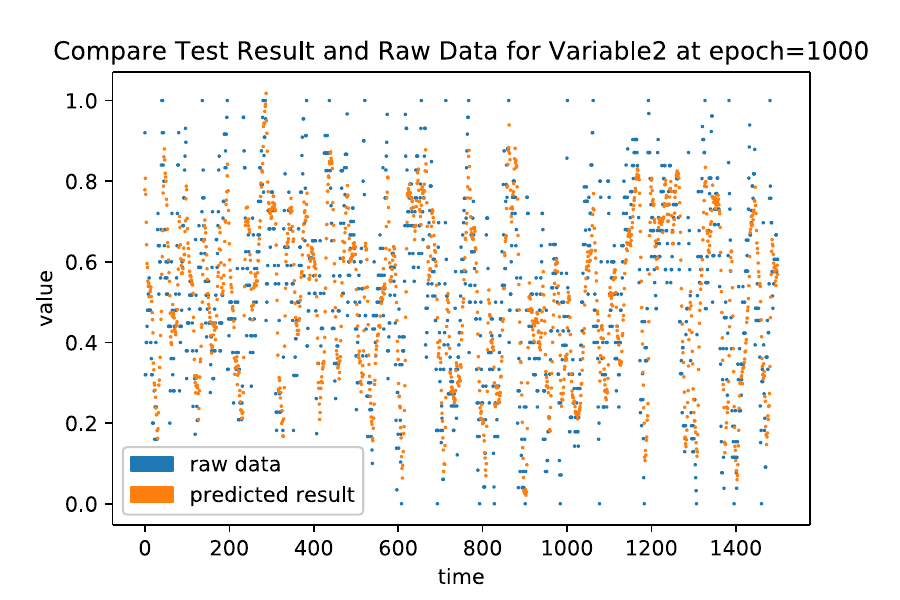}
  \includegraphics[width=0.49\textwidth]{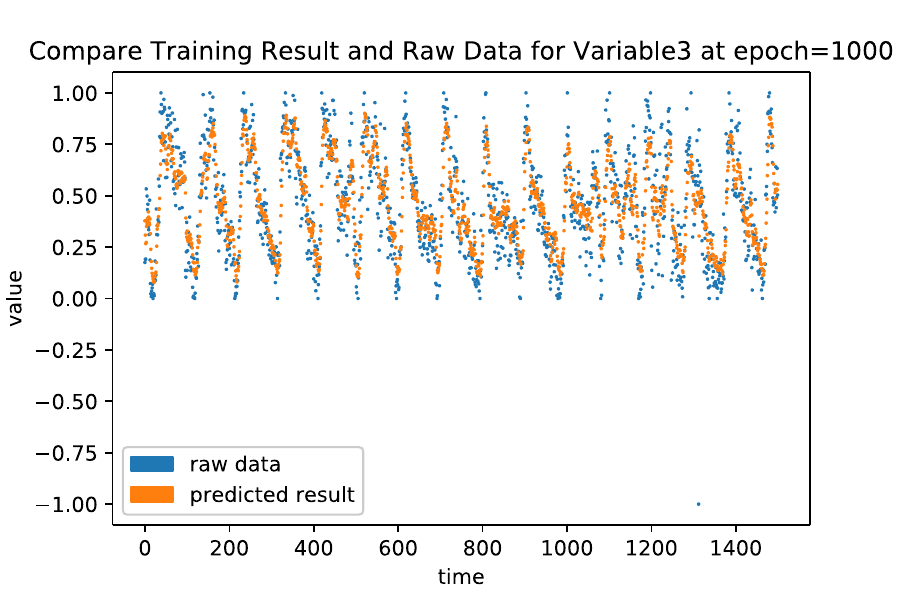}
  \includegraphics[width=0.49\textwidth]{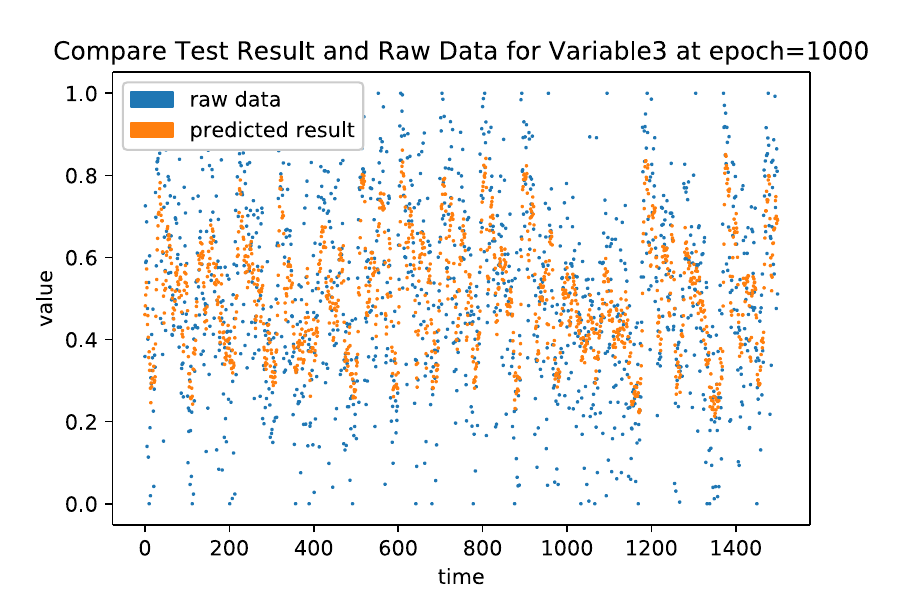}
  \includegraphics[width=0.49\textwidth]{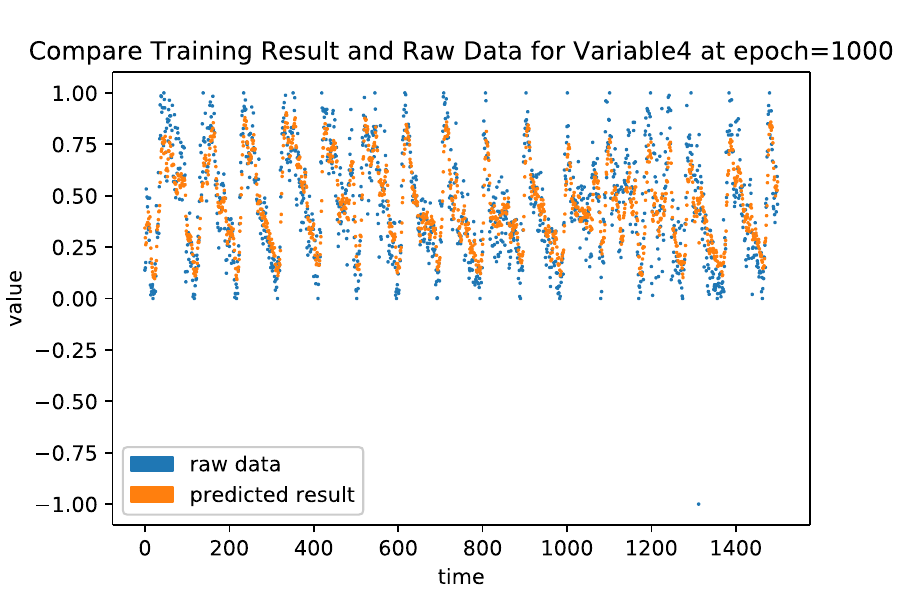}
  \includegraphics[width=0.49\textwidth]{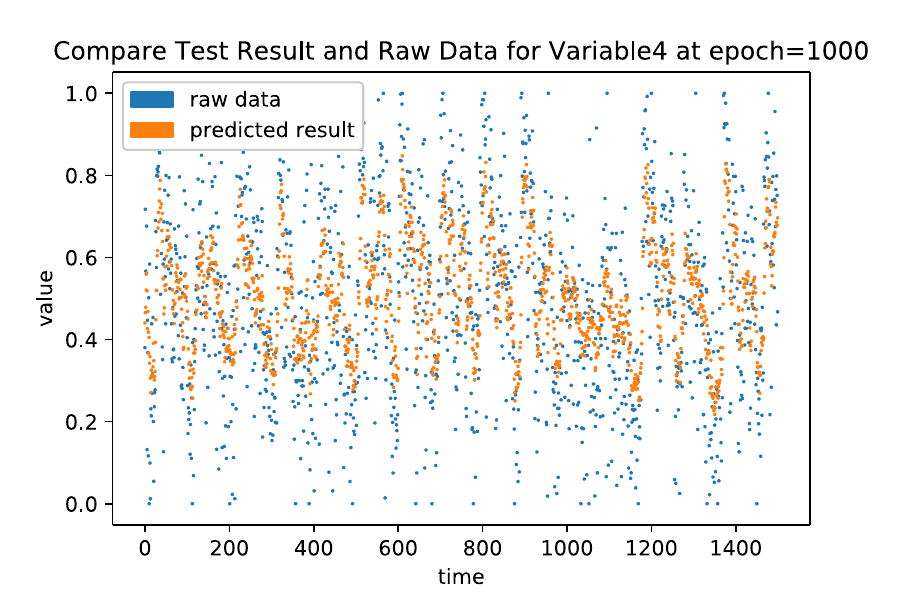}
  \includegraphics[width=0.49\textwidth]{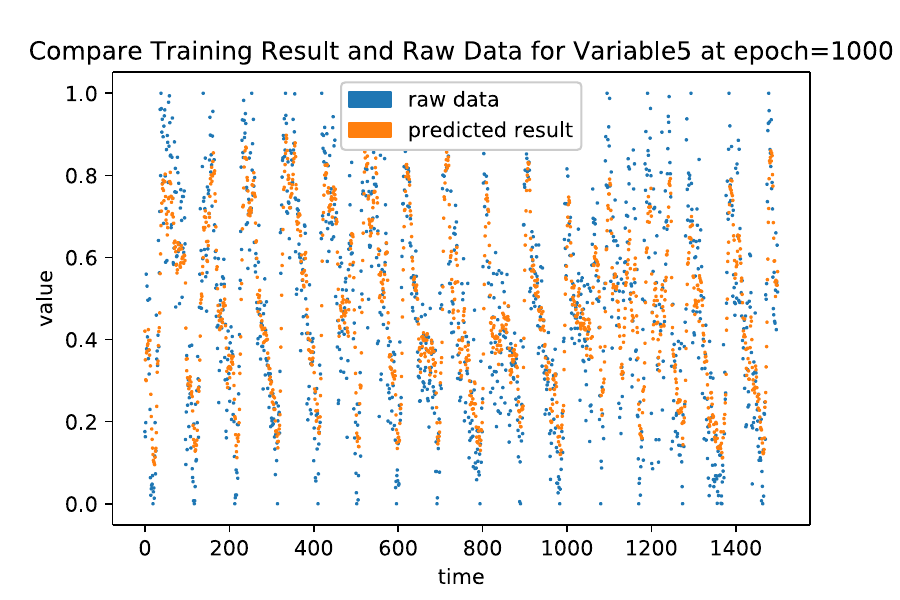}
  \includegraphics[width=0.49\textwidth]{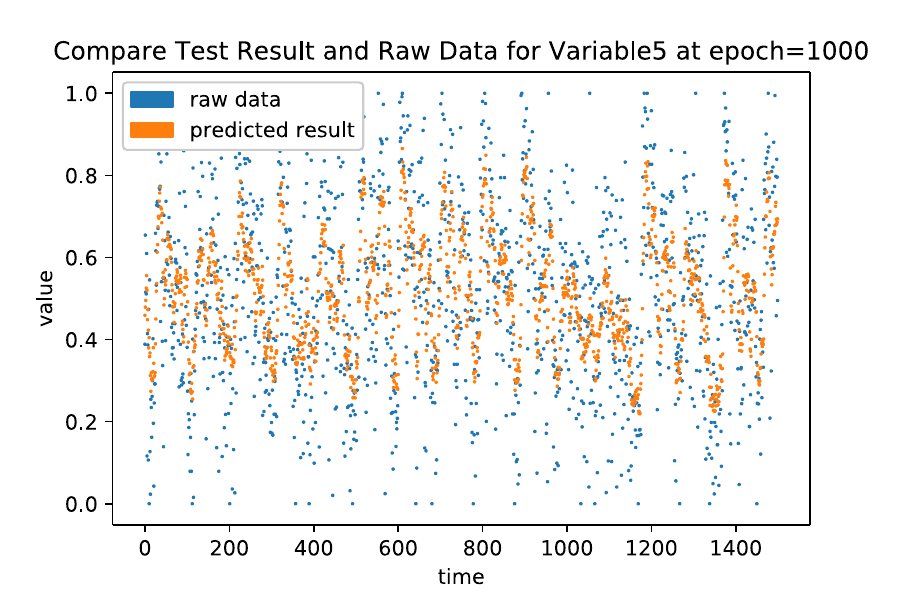}
  \includegraphics[width=0.49\textwidth]{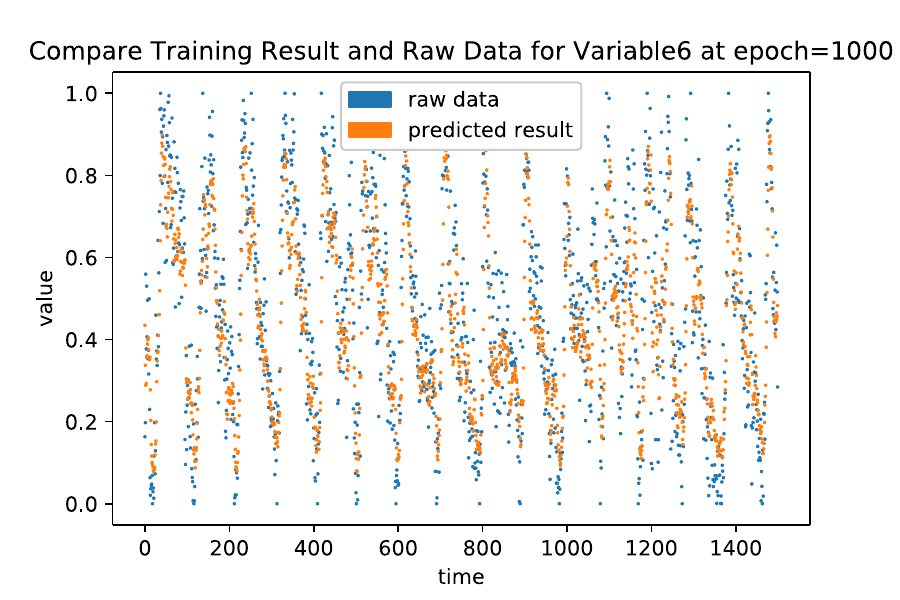}
  \includegraphics[width=0.49\textwidth]{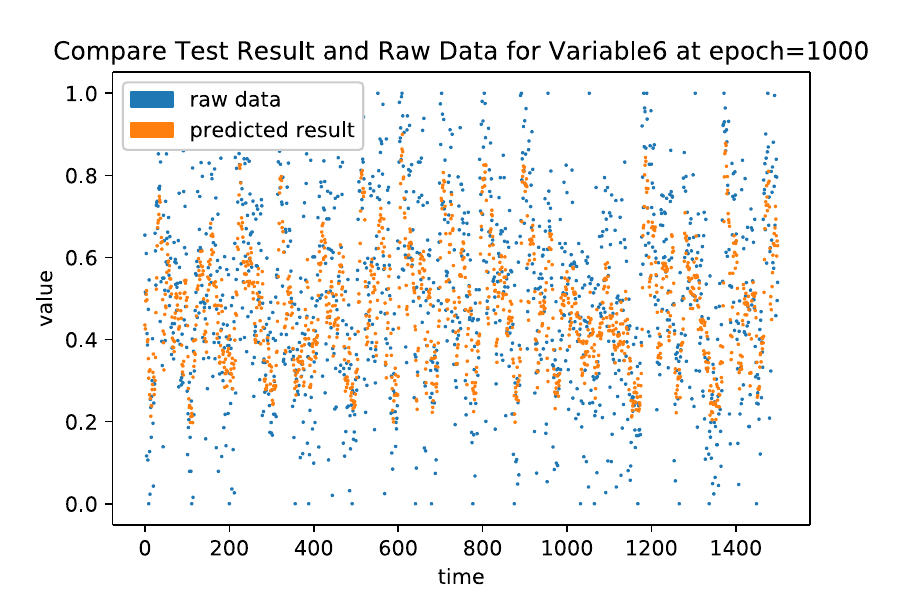}
  \caption{Forecasting results: Industrial enterprises power system dataset}
\end{figure}

\bibliographystyle{unsrt}  




\end{document}